\documentclass[letterpaper]{article} % DO NOT CHANGE THIS
\usepackage{aaai2026}  % DO NOT CHANGE THIS
\usepackage{times}  % DO NOT CHANGE THIS
\usepackage{helvet}  % DO NOT CHANGE THIS
\usepackage{courier}  % DO NOT CHANGE THIS
\usepackage[hyphens]{url}  % DO NOT CHANGE THIS
\usepackage{graphicx} % DO NOT CHANGE THIS
\usepackage{natbib}  % DO NOT CHANGE THIS AND DO NOT ADD ANY OPTIONS TO IT
\usepackage{caption} % DO NOT CHANGE THIS AND DO NOT ADD ANY OPTIONS TO IT
\usepackage{algorithm}
\usepackage{algorithmic}
\usepackage{multirow}  % 必须加载在导言区
\usepackage{array}     % 增强表格对齐
\usepackage{booktabs}
\usepackage{amssymb}   % 包含 amsfonts 的全部功能，并扩展更多符号
\usepackage{amsmath}
\usepackage{enumitem}
\usepackage{tabularx}
\usepackage{makecell}
\usepackage[absolute,overlay]{textpos}

\usepackage{newfloat}
\usepackage{listings}
\DeclareCaptionStyle{ruled}{labelfont=normalfont,labelsep=colon,strut=off} % DO NOT CHANGE THIS
\floatstyle{ruled}
\newfloat{listing}{tb}{lst}{}
\floatname{listing}{Listing}
\title{Consensus-Aligned Neuron Efficient Fine-Tuning Large Language Models for Multi-Domain Machine Translation}
\author{
    Shuting Jiang\textsuperscript{1,2}, Ran Song\textsuperscript{1,2}\footnote{Corresponding author}, Yuxin Huang\textsuperscript{1,2}, Yan Xiang\textsuperscript{1,2}, \\
    Yantuan Xian\textsuperscript{1,2}, Shengxiang Gao\textsuperscript{1,2}, Zhengtao Yu\textsuperscript{1,2}\footnotemark[\value{footnote}]
}

\affiliations{
    \textsuperscript{\rm 1}Faculty of Information Engineering and Automation, \\ Kunming University of Science and Technology, Kunming, China \\ \textsuperscript{\rm 2} Yunnan Key Laboratory of Artificial Intelligence, Kunming, China \\
    \{shuting\_jiang22, song\_ransr, huangyuxin2004\}@163.com, sharonxiang@126.com, \\ xianyt@kust.edu.cn, \{gaoshengxiang.yn, ztyu\}@hotmail.com 
}

\usepackage{bibentry}
\begin{document}

\maketitle

\begin{abstract}
Multi-domain machine translation (MDMT) aims to build a unified model capable of translating content across diverse domains.
Despite the impressive machine translation capabilities demonstrated by large language models (LLMs), domain adaptation still remains a challenge for LLMs. 
Existing MDMT methods such as in-context learning and parameter-efficient fine-tuning often suffer from domain shift, parameter interference and limited generalization. 
In this work, we propose a neuron-efficient fine-tuning framework for MDMT that identifies and updates consensus-aligned neurons within LLMs. 
These neurons are selected by maximizing the mutual information between neuron behavior and domain features, enabling LLMs to capture both generalizable translation patterns and domain-specific nuances.
Our method then fine-tunes LLMs guided by these neurons, effectively mitigating parameter interference and domain-specific overfitting.
Comprehensive experiments on three LLMs across ten German-English and Chinese-English translation domains evidence that our method consistently outperforms strong PEFT baselines on both seen and unseen domains, achieving state-of-the-art performance. 
\end{abstract}

\begin{links}
    \link{Code}{https://github.com/fortunatekiss/CANEFT}
\end{links}

\begin{figure}[t!]
    \centering
    \includegraphics[width=\linewidth]{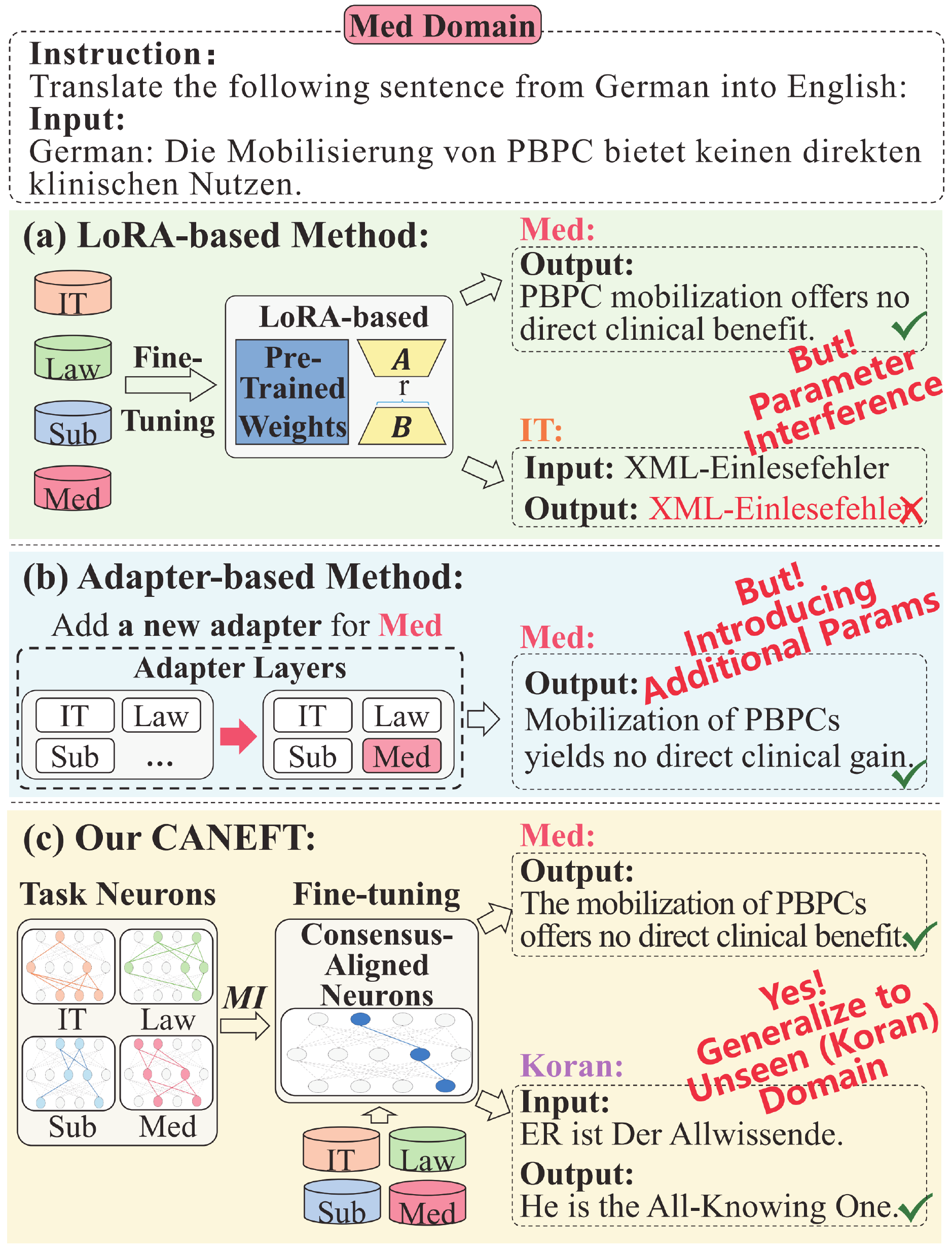}
    \caption{(a) LoRA-based fine-tuning causes parameter interference, while (b) adapter-based methods introduce additional parameters. (c) Our proposed CANEFT addresses these issues by only updating consensus-aligned neurons.}
    \label{fig:Intro}
\end{figure}

% \begin{figure}[t]
%     \centering
%     \resizebox{\linewidth}{0.75\height}{\includegraphics{new_intro_26.pdf}
%     }
%     \caption{(a) LoRA-based fine-tuning causes parameter interference, while (b) adapter-based methods introduce additional parameters. (c) Our proposed CANEFT addresses these issues by only updating consensus-aligned neurons.}
%     \label{fig:Intro}
% \end{figure}

% \begin{textblock*}{0.45\linewidth}(0.53\textwidth,0.5cm)
%     \centering
%     \includegraphics[width=\linewidth]{new_intro_26.pdf}
%     \captionof{figure}{(a) LoRA-based fine-tuning causes parameter interference, while (b) adapter-based methods introduce additional parameters. (c) Our proposed CANEFT addresses these issues by only updating consensus-aligned neurons.}
% \end{textblock*}

\section{Introduction}
Multi-domain machine translation (MDMT) aims to build a unified model capable of accurately translating domain-specific terminology and context across diverse domains such as law, medicine, and subtitles~\cite{pham2021revisiting, saunders2022domain, moslem2023adaptive}. 
Conventional encoder-decoder approaches heavily rely on large amounts of parallel domain data, which are often scarce and costly~\cite{Li_Wang_Yu_2020, saunders2022domain}.
In contrast, Large Language Models (LLMs), pretrained on extensive unlabeled corpora, acquire strong cross-lingual capabilities and show promising performance in general-domain translation~\cite{zhao2023survey, huang2024survey} but still facing challenges when translating domain-specific content~\cite{pang-etal-2025-salute}.
LLMs can notably improve their domain-specific translation capabilities via In-Context Learning (ICL) with a few examples~\cite{ghazvininejad2023dictionary, he-etal-2024-exploring, aycock2024topic, li2025leveraging}. 
However, the performance of ICL depends heavily on the quality of in-domain examples and often degrades for MDMT~\cite{vilar2023prompting}.

Existing studies attempt to address these challenges through parameter-efficient fine-tuning (PEFT) such as LoRA~\cite{alves-etal-2023-steering, zheng2024fine} and adapters~\cite{hu-etal-2023-llm, wu2024llama, eschbach2024exploring}. 
However, LoRA-based methods often face parameter interference. 
As shown in Figure~\ref{fig:Intro}(a), this interference can cause the model to overfit specific domains (Medical) while degrading performance on others (IT).
Adapter-based methods typically introduce separate modules for each domain. 
As shown in Figure~\ref{fig:Intro}(b), these methods increase training and memory costs as the number of domain grows, and lacks generalization to unseen domains.
In summary, these limitations motivate a central question: \textit{Can we design a robust PEFT method for multi-domain machine translation without introducing additional parameters?}

Recent research has investigated neuron behavior in LLMs for tasks such as multilingual machine translation~\cite{zhu2024landermt}, arithmetic reasoning~\cite{hersche2024towards, rai2024investigation} and knowledge editing~\cite{jiang2024neuron, li2024knowledge}. 
These studies indicate that neuron subsets spontaneously encode language- or task-specific functions, suggesting the potential of neuron-based approaches to enhance both performance and efficiency.
Inspired by these findings, we delve into the correlation between neurons behavior and MDMT task, aiming to disentangle MDMT capabilities from a neuronal perspective in LLMs.
Previous studies on task-specific neuron selection and fine-tuning have primarily focused on identifying neurons based on activation patterns or gradient variations~\cite{song2024does,tang2024language}.
However, these methods frequently select neurons either irrelevant to MDMT or overly specialized for single domains, which hinders generalization and causes domain-specific overfitting.

Intriguingly, recent neuroscience research shows that consensus-building dialogues enhance neural alignment among group members, with synchronized brain activity even generalizing to novel, unseen stimuli~\cite{Sievers2024consensus}.
Motivated by this, we hypothesize that within LLMs, certain neurons might consistently encode knowledge relevant across multiple domains. 
We term these neurons \textit{consensus-aligned neurons}.
As illustrated in Figure~\ref{fig:Intro}(c), fine-tuning on consensus-aligned neurons achieves enhanced translation performance and generalization across multiple domains, contrasting with strategies that isolating domain-specific neurons.
Consequently, accurately identifying the consensus-aligned neurons is a crucial prerequisite for improving MDMT performance.

In this work, we propose \textbf{C}onsensus-\textbf{A}ligned \textbf{N}euron \textbf{E}fficient \textbf{F}ine-\textbf{T}uning (CANEFT) for MDMT.
% \footnote{The code for this paper is available at https://github.com/fortunatekiss/CANEFT.}
This framework identifies and leverages consensus-aligned neurons to improve translation performance on both seen and unseen domains.
Specifically, we first detect MDMT task-related neurons through activation-gradient analysis during inference.
We then compute the mutual information (MI) between each task neuron and each domain to pinpoint consensus-aligned neurons. 
Finally, our approach significantly enhance LLM multi-domain translation performance by masking irrelevant neurons and fine-tuning on the identified consensus-aligned neurons.
We conducted extensive experiments and analysis for German-English and Chinese-English translation across 10 domains.
The results show that our method surpasses the full fine-tuning baseline by an average of 1.3 BLEU on De$\Rightarrow$En and 1.4 BLEU on Zh$\Rightarrow$En, also demonstrating strong generalization to unseen domains.

The main contributions of this work are as follows:
\setlength{\leftmargini}{0.3cm}
\begin{itemize}
    \item We introduce consensus-aligned neurons critical for MDMT through MI-based strategy. These neurons effectively mitigate parameter interference and reduce the need for extensive fine-tuning domain data.
    \item We propose a neuron-efficient fine-tuning framework for MDMT, which selectively updates multi-domain consensus-aligned neurons to enhance both translation quality and cross-domain generalization.
    \item We validated our method on 3 instruction-tuned LLMs across 10 domain translation tasks in German-English and Chinese-English. 
    Our method achieves an average BLEU improvement of 1.3 on De$\Rightarrow$En and 1.4 on Zh$\Rightarrow$En over the best-performing baseline, and demonstrates robust generalization to unseen domains.
\end{itemize}

\section{Related Work}
\subsection{Domain Machine Translation}
Recent research on domain-specific machine translation with LLMs has explored both inference-time adaptation and fine-tuning.
\citet{aycock2024topic} propose a topic-guided demonstration retrieval method to enhance translation performance of ICL without fine-tuning.
\citet{li2025leveraging} compare retrieval- and generation-based domain prompting, showing retrieval provides better grounding while generation enables flexibility.
\citet{hu2024large} build a MDMT benchmark and enhance cross-domain performance via domain-aware chain-of-thought fine-tuning.
\citet{dragft} introduce dictionary- and retrieval-augmented fine-tuning to bridge terminology gaps in domain translation.

\subsection{Neuron Analysis in LLMs}
Recent studies have revealed that neurons in LLMs exhibit modular behavior, with certain neurons responsible for specific languages~\cite{tan2024neuron, tang2024language, zhu2024landermt}, tasks~\cite{song2024does, leng2025towards}, or knowledge~\cite{dai2022knowledge, chen2024journey, niudoes, mao-etal-2025-multilingual}. This intrinsic structure has motivated a growing works on neuron-level analysis~\cite{voita2024neurons} and selective fine-tuning~\cite{xu2025let}.
\citet{tan2024neuron} show that neuron activations correlate with language typological proximity, while \citet{tang2024language} introduce LAPE to identify language-specific neurons and control output via targeted activation.
\citet{zhu2024landermt} achieve strong multilingual translation performance by routing updates through language-specific and general neurons.
For tasks, \citet{song2024does} and \citet{leng2025towards} identify task-specific neurons using activation patterns and gradient attribution, respectively.
In terms of factual knowledge, \citet{dai2022knowledge} define "knowledge neurons" that activate for specific facts.
\citet{chen2024journey} shedding light on the mechanisms of cross-lingual factual knowledge storage in LLM neurons.

Unlike previous work, we identify consensus-aligned neurons for MDMT by measuring mutual information with domain context, and propose a neuron-efficient fine-tuning framework that enhances cross-domain translation while mitigating PEFT limitations.

\section{Methodology}
In this section, we exhaustively introduce our neuron-efficient fine-tuning framework, designed to enhance LLMs performance and generalization on MDMT through identifying and selectively updating multi-domain consensus-aligned neurons. The framework consists of three key steps:
\textit{(i) \textbf{MDMT task-relevant neuron identification}} identifies MDMT task related neurons in feed-forward network (FFN) through calculating neuron activation sensitivity and gradient magnitude;
\textit{(ii) \textbf{MI-based multi-domain consensus-aligned neuron selection}} measures MI between importance of these task-relevant neurons and domain features, then select a small critical subset as MDMT consensus-aligned neurons that capture cross-domain translation knowledge as well as domain-specific nuances;
\textit{(iii) \textbf{Neuron-efficient fine-tuning}} only updates the MDMT consensus-aligned neurons using domain translation examples, enabling multi-domain adaptation and robust generalization.

\subsection{MDMT Task-Relevant Neuron Identification}
\label{sec:MDTN}
Neurons that consistently exhibit strong gradient and activation responses to task-specific inputs are likely to encode MDMT-relevant features. To identify such neurons, we compute gradient-activation importance scores for each neuron in the FFN layers of an LLM $f_\theta$ during MDMT inference. 

For each domain $d$, we utilize parallel data 
$D_d=(\mathcal{T}^d, \mathbf{x}^d, \mathbf{y}^d)$, where 
$\mathbf{x}^d=\{x^d_1,...,x^d_K\}$ is the source sequence and $\mathbf{y}^d=\{y^d_1,...,y^d_T\}$ is the target sequence. $\mathcal{T}^d$ is the domain-specific instruction designed to inform the LLM of the domain context. 
For example, a De$\Rightarrow$En IT domain instruction is: "\textit{You are a translation specialist who specializes in translating texts from German to English in the IT domain. Translate the following content into English and only reply to the translated sentence without line breaks or special symbols.}"

Inspired by studies on importance-based neuron selection for multilingual machine translation~\cite{xie2021importance}, we assess a neuron's importance by multiplying its activation during the forward pass by the loss gradient with respect to that activation during the backward pass.
Neurons with high importance scores are identified as strongly associated with the MT task for a given domain $d$.

Specifically, let $A_{l,j}^{(d)}$ denote the activation of the $j$-th neuron in the $l$-th FFN layer when processing an input $\mathbf{x}^d$ from domain $d$, and $G_{l,j}^{(d)}$ be the  gradient of the loss with respect to that activation:
\begin{equation}
G_{l,j}^{(d)} = \frac{\partial \mathcal{L}^{(d)}(\mathbf{x}^d, \mathbf{y}^d)}{\partial h_{l,j}^{(d)}},
\end{equation}
where $h_{l,j}^{(d)}$ is the output of the $j$-th neuron in the $l$-th FFN layer for domain $d$.
And the token-level cross-entropy loss $\mathcal{L}^{(d)}$ between the model’s prediction and the reference translation $\mathbf{y}^d$ is:
\begin{equation}
\mathcal{L}^{(d)}(\mathbf{x}^d, \mathbf{y}^d) = -\sum_{t=1}^{T} \log p_\theta(y_t^d \mid y_{<t}^d, \mathbf{x}^d, \mathcal{T}^d).
\end{equation}
 % and $\mathcal{L}^{(d)}$ is the token-level cross-entropy loss between the model’s prediction and the reference translation $\mathbf{y}^d$.
The neuron importance score for $d$ is then computed as:
\begin{equation}
I_{l,j}^{(d)} = \mathbb{E}{(\mathbf{x}^d, \mathbf{y}^d) \sim D_d} \left[ \left| A_{l,j}^{(d)} \cdot G_{l,j}^{(d)} \right| \right].
\end{equation}

To further prove the proposed gradient-based activation importance metric, we approximate a neuron's contribution by evaluating the change in loss when the neuron is removed. 
Thus, we apply a first-order Taylor Expansion~\cite{molchanov2017pruning} to estimate the impact of ablating individual neurons.
Let $H^{(d)}$ represent neurons in layer $l$ excluding the $j$-th neuron.
Assuming the output of each neuron contributes independently to the loss, the change in loss due to removing the $j$-th neuron can be expressed as:
\begin{align}
\left|\Delta \mathcal{L}^{(d)}(h_{l,j}^{(d)})\right| 
&= \left| \mathcal{L}^{(d)}(H^{(d)}, h_{l,j}^{(d)} = 0) \right. \nonumber \\
&\quad \left. - \mathcal{L}^{(d)}(H^{(d)}, h_{l,j}^{(d)}) \right|, 
\end{align}
where $\mathcal{L}^{(d)}(H^{(d)}, h_{l,j}^{(d)} = 0)$ is the loss on domain $d$ when the $j$-th neuron is removed, and $\mathcal{L}^{(d)}(H^{(d)}, h_{l,j}^{(d)})$ is the loss when it is retained. 
Then, applying a first-order Taylor approximation, this change in the loss for domain $d$ can be estimated as:
\begin{equation}
\begin{aligned}
\mathcal{L}^{(d)}(H^{(d)}, h_{l,j}^{(d)})
&=\mathcal{L}^{(d)}(H^{(d)},h_{l,j}^{(d)}=0)\\
&+\frac{\partial\mathcal{L}^{(d)}(H^{(d)},h_{l,j}^{(d)})}{\partial{h_{l,j}^{(d)}}}+R_1(h_{l,j}^{(d)}),
\end{aligned}
\end{equation}
where $R_1(h_{l,j})$ is the Lagrange remainder term associated with the approximation for domain $d$:
\begin{equation}
    R_1(h_{l,j}^{(d)})=\frac{\partial^2\mathcal{L}^{(d)}(H^{(d)},h_{l,j}^{(d)})}{\partial^2\delta{h_{l,j}^{(d)}}}(h_{l,j}^{(d)})^2,
\end{equation}
where $\delta \in (0,1)$.
The first derivative of the loss function with respect to the neuron's output for domain $d$ tends to become constant. Consequently, during the final stages of training for domain $d$, the second-order term approaches zero.
Therefore, by neglecting the remainder term, the importance evaluation function for a neuron with respect to domain $d$ can be approximated as:
\begin{equation}
    |\Delta\mathcal{L}^{(d)}(h_{l,j}^{(d)})| \approx \left|\frac{\partial \mathcal{L}^{(d)}(H^{(d)}, h_{l,j}^{(d)})}{\partial h_{l,j}^{(d)}} \cdot h_{l,j}^{(d)}\right|.
\end{equation}

Thus, the importance score $I_{l,j}^{(d)}$ serves as an approximation to the contribution of neuron $j$ in layer $l$ to the MDMT loss in domain $d$:
\begin{equation}
\begin{aligned}
I_{l,j}^{(d)}
&= \mathbb{E}{(\mathbf{x}^d, \mathbf{y}^d) \sim D_d} \left[ \left| A_{l,j}^{(d)} \cdot G_{l,j}^{(d)} \right| \right] \\
&\approx \left| \frac{\partial \mathcal{L}^{(d)}(H^{(d)}, h_{l,j}^{(d)})}{\partial h_{l,j}^{(d)}} \cdot h_{l,j}^{(d)} \right|,
\end{aligned}
\end{equation}
which serves as the foundation for multi-domain consensus-aligned neuron selection, as detailed in the next subsection. 

\subsection{MI-based Multi-Domain Consensus-Aligned Neuron Selection} \label{sec:MDCA}
After computing MDMT task-relevant importance scores $I^{(d)}_{l,j}$, we identify consensus-aligned neurons that consistently exhibit high relevance to translation across all domains. 
These neurons capture domain-invariant translation mechanisms while retaining sensitivity to domain-specific nuances, thereby enabling generalization to unseen domains and efficient fine-tuning for seen ones.

To quantify the relationship between a neuron's importance and the domain identity, we employ MI measurement. 
MI can precisely quantify the statistical dependencies between variables, which enables the assessment of whether neurons are multi-domain consensus-aligned.
Specifically, we first discretize the continuous importance scores into a set of fixed bins indexed by $i$ to facilitate probability estimation over domain feature.
For the $j$-th neuron in the $l$-th FFN layer, we then estimate its MI with the domain label $d$ as follows:
\begin{equation}
    \begin{aligned}
        &\text{MI}_{l,j}
        = \sum_i \sum_{d \in D} \! H(I^{(d)}_{l,j}) \! + \! H(d) \! - \! H(I^{(d)}_{l,j}, d) \\
        &= \sum_{i} \sum_{d \in D} p(I_{l,j}^{(d)} \! = \! i, d) \log \Bigl( \frac{p(I_{l,j}^{(d)} \! = \! i, d)}{p(I_{l,j}^{(d)} \! = \! i) p(d)} \Bigr),
    \end{aligned}
\end{equation}
where term $p(I_{l,j}^{(d)} = i, d)$ denotes the joint probability that the neuron's importance score $I_{l,j}^{(d)}$ falls into importance bin $i$ for a data sample belonging to domain $d$. 
This captures the joint behavior between neuron importance and domain identity, reflecting how frequently a neuron exhibits a particular level of importance within a specific domain. 
The marginal probability $p(I_{l,j}^{(d)} = i)$ represents the overall distribution of the neuron's importance score across all domains, indicating how frequently its score falls into bin $i$ regardless of the domain. The term $p(d)$ is the marginal probability of encountering a data sample from domain $d$.

To select our multi-domain consensus-aligned neurons $\mathcal{N}_{\text{MDCA}}$, we avoid selecting neurons with high MI for individual domains, as this can introduce domain-specific noise and lead to overfitting. Instead, we identify neurons that consistently exhibit high MI across all domains:
\begin{equation}
    \mathcal{N}_{\text{MDCA}} = \left\{ (l, j) \mid \min \text{MI}_{l,j} \geq \gamma \right\},
    \label{eq:8}
\end{equation}
where $\gamma$ is a threshold, and ensures that a neuron is selected only if its MI with the domain label is at least $\gamma$ in every domain $d\in D$. This allows us to identify neurons that are robustly aligned with multi-domain characteristics across the entire set of domains $D$.
\subsection{Neuron-Efficient Fine-Tuning} 
We propose a Neuron-Efficient Fine-Tuning strategy that only updates the multi-domain consensus-aligned neurons.

Formally, let $\mathcal{M}_\theta$ denotes a LLM with parameters $\theta$, we freeze all parameters except those directly associated with neurons in $\mathcal{N}_\text{MDCA}$. 
The FFN layer contains three modules, up projection, down projection and gate projection. For each module $m$, let $W_m \in \mathbb{R}^{in \times out}$ denotes the weight matrix, where $in$ is the input dimension to the module and $out$ is the output dimension.
The number of neurons is identical to $out$.
To enable selective gradient updates, we construct a binary mask $M \in \mathbb{R}^{in \times out}$, such that:
\begin{equation}
    M_{:i}=
    \begin{cases}
        1, & \quad \text{if } i \in \mathcal{N}_{\text{MDCA}}\\
        0, & \quad \text{otherwise}.
    \end{cases}
\end{equation}

This mask is applied during the backward pass to suppress gradients for all non-selected neurons. Specifically, the gradient update for each weight matrix $W_m$ is modified as:
\begin{equation}
    \nabla{W_m} \leftarrow \nabla{W_m} \odot M,
\end{equation}
where $\odot$ denotes element-wise multiplication. 
Consequently, only the parameters corresponding to the multi-domain consensus-aligned neurons in $\mathcal{N}_\text{MDCA}$ are updated.

\begin{table*}[ht]
    \centering
    \small{  
    \setlength{\tabcolsep}{1mm}
    \resizebox{\textwidth}{!}{
    \begin{tabular}{llcccccccccccccccccccccccc}
        \toprule
        & & \multicolumn{12}{c}{\textbf{De $\Rightarrow$ En}} & \multicolumn{12}{c}{\textbf{Zh $\Rightarrow$ En}} \\
        \cmidrule(lr){3-14} \cmidrule(lr){15-26}
        & & \multicolumn{8}{c}{\textbf{Seen}} & \multicolumn{2}{c}{\textbf{Unseen}} & & &\multicolumn{6}{c}{\textbf{Seen}} & \multicolumn{4}{c}{\textbf{Unseen}} \\
        \cmidrule(lr){3-10} \cmidrule(lr){11-12} \cmidrule(lr){15-20} \cmidrule(lr){21-24}
        \textbf{Methods} & \textbf{TP/AP} & \multicolumn{2}{c}{\textbf{IT}} & \multicolumn{2}{c}{\textbf{Law}} & \multicolumn{2}{c}{\textbf{Med}} & \multicolumn{2}{c}{\textbf{Sub}} & \multicolumn{2}{c}{\textbf{Kor}} & \multicolumn{2}{c}{\textbf{Avg}} & \multicolumn{2}{c}{\textbf{Edu}} & \multicolumn{2}{c}{\textbf{Spo}} & \multicolumn{2}{c}{\textbf{The}} & \multicolumn{2}{c}{\textbf{Sci}} & \multicolumn{2}{c}{\textbf{Blog}} & \multicolumn{2}{c}{\textbf{Avg}}\\
        \cmidrule(lr){3-4} \cmidrule(lr){5-6} \cmidrule(lr){7-8} \cmidrule(lr){9-10} \cmidrule(lr){11-12} \cmidrule(lr){13-14} \cmidrule(lr){15-16} \cmidrule(lr){17-18} \cmidrule(lr){19-20} \cmidrule(lr){21-22} \cmidrule(lr){23-24} \cmidrule(lr){25-26}
        & & B & C & B & C & B & C & B & C & B & C & B & C & B & C & B & C & B & C & B & C & B & C & B & C \\
        \midrule
        \multicolumn{26}{c}{\textbf{LLaMA2-7B-Chat}} \\
        Base & 0/0 & 38.7 & 81.8 & 30.0 & 82.8 & 44.1 & 87.4 & 35.5 & 85.0 & 16.0 & 73.1 & 32.8 & 82.0 & 25.7 & 83.9 & 26.7 & \underline{83.8} & 18.1 & 76.9 & 11.9 & 73.8 & 14.4 & 73.4 & 19.3 & 78.3 \\
        Full & 7k/0 & 43.7 & 87.8 & \textbf{47.9} & 82.8 & 47.1 & 85.5 & 35.3 & 81.1 & \textbf{20.5} & \underline{73.8} & \underline{38.9} & 82.2 & 28.2 & 79.9 & \underline{29.6} & 78.9 & \underline{22.7} & 78.4 & 14.9 & 72.2 & 18.2 & 73.9 & 22.7 & 76.6 \\
        LoRA & 20/20 & \underline{46.7} & 87.4 & 35.8 & 80.1 & \underline{49.9} & 87.7 & 33.5 & 82.5 & 15.3 & 69.6 & 36.2 & 81.4 & 28.2 & 82.0 & 28.6 & 82.1 & 21.6 & 77.1 & 16.3 & 74.3 & 15.3 & 73.8 & 22.0 & 77.8 \\
        L-MLP & 12/12 & 46.4 & 86.9 & 40.8 & 82.5 & 49.7 & \underline{88.0} & 33.7 & 82.5 & 14.4 & 70.2 & 37 & 82.0 & 27.3 & 82.7 & 28.3 & 82.4 & 21.6 & 77.7 & 15.5 & 74.2 & 18.7 & 76.7 & 22.2 & 78.7 \\
        DoRA & 22/22 & 45.9 & 87.8 & 42.4 & \underline{84.7} & 46.7 & 87.7 & 37.5 & 84.3 & 17.6 & 73.2 & 38.0 & \underline{83.5} & 28.7 & 81.9 & 29.0 & 82.2 & 21.6 & 77.0 & \textbf{16.6} & 75.2 & \underline{19.2} & 76.8 & \underline{23.0} & 78.6 \\
        Pro & 405/405 & 40.4 & 83.9 & \underline{46.0} & \textbf{84.9} & 43.5 & 87.2 & 31.3 & 81.3 & 16.4 & 72.4 & 35.5 & 81.9 & \textbf{29.3} & \underline{84.1} & 28.9 & 83.0 & \textbf{23.5} &\textbf{80.0} & 15.1 & \underline{75.3} & 18.0 & \underline{77.2} & 22.9 & \underline{79.9} \\
        LAPE & 91/0 & 39.3 & 80.6 & 30.4 & 80.5 & 40.5 & 85.8 & 29.0 & 76.2 & 15.5 & 73.1 & 30.9 & 79.2 & 21.4 & 82.8 & 22.0 & 82.0 & 15.4 & 76.5 & 13.2 & 73.9 & 14.5 & 73.5 & 17.3 & 77.7 \\
        RCN & 91/0 & 25.5 & 84.1 & 30.8 & 82.0 & 33.6 & 85.8 & \underline{40.4} & \underline{86.1} & 14.6 & 72.7 & 28.9 & 82.1 & 25.1 & 76.9 & 25.0 & 74.0 & 17.7 & 75.1 & 10.7 & 70.1 & 12.2 & 70.8 & 18.1 & 73.3 \\
        CANEFT & 91/0 & \textbf{48.8} & \textbf{88.7} & 43.8 & 84.4 &\textbf{50.0} & \textbf{88.3} & \textbf{43.7} & \textbf{86.8} & \underline{19.2} & \textbf{74.1} & \textbf{40.5} & \textbf{84.5} & \underline{28.9} & \textbf{84.4} & \textbf{30.3} & \textbf{84.2} & 22.6 & \underline{78.5} & \underline{16.4} & \textbf{75.7} & \textbf{20.3} & \textbf{77.8} & \textbf{23.7} & \textbf{80.1} \\
        \midrule
        \multicolumn{26}{c}{\textbf{LLaMA3.1-8B-Instruct}} \\
        Base & 0/0 & \underline{48.1} & \underline{87.8} & 38.3 & 84.4 & 46.4 & 87.4 & \underline{44.4} & \underline{87.4} & 19.3 & 75.5 & 39.3 & 84.5 & 30.7 & 85.2 & 31.1 & 86.0 & 21.6 & 80.6 & 15.4 & 77.4 & 17.1 & 79.1 & 23.1 & 81.6 \\
        Full & 8k/0 & 47.2 & 84.3 & \textbf{56.3} & \textbf{88.0} & \underline{51.4} & \underline{88.7} & 42.1 & 87.2 & \textbf{21.0} & 75.5 & \underline{43.6} & \underline{84.7} & \underline{31.3} & 85.3 & \underline{33.4} & 84.6 & 25.5 & \textbf{81.6} & 16.2 & \textbf{78.1} & 19.4 & \underline{79.2} & \underline{25.1} & \underline{81.7} \\
        LoRA & 42/42 & 45.8 & 85.9 & 45.8 & 85.3 & 50.8 & 88.5 & 34.6 & 82.4 & 18.8 & 72.2 & 39.1 & 82.8 & 30.4 & 84.4 & 30.6 & 83.5 & 24.3 & 80.2 & \underline{17.9} & 77.1 & \underline{21.7} & 78.0 & 24.9 & 80.6 \\
        L-MLP & 29/29 & 43.2 & 85.4 & 48.4 & 85.5 & 47.5 & 87.9 & 34.9 & 82.9 & 18.9 & 73.8 & 38.5 & 83.1 & 30.3 & 84.5 & 30.2 & 83.5 & \textbf{25.8} & 80.6 & 16.6 & 77.1 & 21.5 & 78.1 & 24.8 & 80.7 \\
        DoRA & 44/44 & 44.3 & 85.5 & 50.0 & 85.9 & 49.3 & 88.4 & 35.7 & 83.1 & 19.5 & 73.9 & 39.7 & 83.3 & 30.4 & 84.4 & 30.5 & 83.5 & 24.5 & 80.1 & 16.7 & 76.1 & 21.1 & 78.2 & 24.6 & 80.4 \\
        Pro & 436/436 & 44.7 & 85.5 & 50.9 & 86.1 & 49.8 & 88.4 & 34.8 & 82.6 & 20.2 & 74.2 & 40.0 & 83.3 & 29.3 & 84.1 & 28.9 & 83.0 & 23.5 & 80.0 & 17.0 & 77.5 & 20.3 & 78.4 & 23.8 & 80.6 \\
        LAPE & 146/0 & 46.2 & 87.6 & 42.9 & 85.3 & 44.9 & 87.6 & 41.8 & 87.3 & 18.1 & 73.9 & 38.7 & 84.3 & 28.8 & \underline{86.3} & 28.6 & \underline{86.2} & 21.0 & 80.5 & 15.9 & 76.4 & 20.1 & 77.9 & 22.8 & 81.4 \\
        RCN & 146/0 & 41.7 & 83.8 & 40.6 & 84.5 & 47.7 & 87.5 & 42.6 & 86.2 & 20.6 & \underline{75.8} & 38.6 & 83.5 & 31.2 & 85.9 & 30.9 & 85.7 & 22.2 & 81.0 & 14.2 & 75.8 & 15.0 & 76.6 & 22.7 & 81.0 \\
        CANEFT & 146/0 & \textbf{54.8} & \textbf{90.7} & \underline{50.9} & \underline{87.1} & \textbf{52.3} & \textbf{90.3} & \textbf{46.3} & \textbf{88.0} & \underline{20.9} & \textbf{75.9} & \textbf{45.0} & \textbf{86.4} & \textbf{34.1} & \textbf{86.5} & \textbf{34.8} & \textbf{86.5} & \underline{25.6} & \underline{81.5} & \textbf{18.9} & \underline{77.7} & \textbf{22.1} & \textbf{79.4} & \textbf{27.1} & \textbf{82.3} \\
        \midrule
        \multicolumn{26}{c}{\textbf{Qwen2.5-7B-Instruct}} \\
        Base & 0/0 &  \underline{54.3} & \underline{90.5} & 42.6 & 86.2 & 50.3 & 89.5 & \underline{44.0} & 87.8 & 19.4 & 75.1 & 42.1 & \underline{85.8} & 33.9 & 86.4 & 34.2 & 86.4 & 23.4 & 81.1 & 18.3 & 77.9 & 19.3 & 79.1 & 25.8 & 82.1 \\
        Full & 7k/0 & 53.2 & 87.8 & \textbf{53.5} & 85.9 & \textbf{53.1} & 89.9 & 42.2 & 87.1 & \underline{21.2} & \underline{75.6} & \underline{44.6} & 85.2 & \underline{35.1} & \textbf{87.2} & \underline{36.8} & \underline{87.3} & 26.6 & 81.3 & 20.6 & \underline{78.3} & 21.5 & 79.2 & \underline{28.1} & \underline{82.6} \\
        LoRA & 21/21 & 46.0 & 85.9 & 47.1 & 85.5 & 47.4 & 87.8 & 36.1 & 83.4 & 18.6 & 74.4 & 39.0 & 83.4 & 31.7 & 84.9 & 31.8 & 84.1 & 25.8 & 80.6 & \underline{21.1} & 78.0 & 22.4 & 78.8 & 26.5 & 81.2 \\
        L-MLP & 12/12 & 45.9 & 85.9 & 47.0 & 85.4 & 47.6 & 87.8 & 35.8 & 83.3 & 18.6 & 74.3 & 38.9 & 83.3 & 31.2 & 85.0 & 31.5 & 83.9 & 25.8 & 80.7 & 21.1 & 77.9 & \underline{22.7} & 78.8 & 26.4 & 81.2 \\
        DoRA & 22/22 & 45.4 & 85.8 & 47.2 & 85.4 & 47.9 & 87.8 & 36.1 & 83.3 & 18.8 & 74.4 & 39.0 & 83.3 & 31.5 & 84.9 & 31.8 & 84.0 & 25.7 & 80.6 & 20.9 & 77.9 & 21.8 & 78.8 & 26.3 & 81.2 \\
        Pro & 466/466 & 45.7 & 85.7 & \underline{50.6} & \underline{86.7} & 50.0 & 88.4 & 35.0 & 82.9 & 19.7 & 74.2 & 40.2 & 83.5 & 31.0 & 84.8 & 31.3 & 83.7 & 25.6 & 80.6 & 18.5 & 78.1 & 21.0 & 79.1 & 25.4 & 81.2 \\
        LAPE & 127/0 & 52.7 & 89.7 & 45.0 & 85.8 & 49.0 & \underline{89.9} & 42.9 & \underline{88.0} & 17.3 & 74.7 & 41.3 & 85.6 & 32.9 & 87.0 & 32.7 & 87.2 & 23.9 & 81.3 & 17.8 & 77.5 & 19.4 & \underline{79.3} & 25.3 & 82.4 \\
        RCN & 127/0 & 50.5 & 89.1 & 45.1 & 86.3 & 50.3 & 88.9 & 40.9 & 85.4 & 17.6 & 73.9 & 40.8 & 84.7 & 33.7 & 86.1 & 35.0 & 86.1 & \underline{26.9} & \underline{82.1} & 17.5 & 76.2 & 18.6 & 78.2 & 26.3 & 81.7 \\
        CANEFT & 127/0 & \textbf{56.5} & \textbf{91.4} & 49.8 & \textbf{87.2} & \underline{52.3} & \textbf{90.1} & \textbf{47.2} & \textbf{88.3} & \textbf{21.5} & \textbf{76.1} & \textbf{45.5} & \textbf{86.6} & \textbf{36.9} & \underline{87.1} & \textbf{38.0} & \textbf{87.4} & \textbf{27.2} & \textbf{82.2} & \textbf{22.7} & \textbf{78.9} & \textbf{23.2} & \textbf{79.9} & \textbf{29.6} & \textbf{83.1} \\
        \bottomrule
    \end{tabular}
    }
    }
    \caption{Translation performance of 3 distinct models across IT, Law, Medical (Med), Subtitles (Sub), Koran (Kor), Education (Edu), Spoken (Spo), Thesis (The), Science (Sci), and Microblog (Blog) domains, evaluated using BLEU (B) and COMET (C) scores. "TP" denotes Trainable Parameters and "AP" refers to Additional Parameters, both measured in \textbf{millions}. The \textbf{best} results for each metric and domain are bolded, and the \underline{second-best} are underlined.}
    \label{tab:translation-results}
\end{table*}

\section{Experiments}
\subsection{Experimental Setups}
\textbf{Dataset} 
We conducted experiments in 2 translation directions: German-English (De$\Rightarrow$En) and Chinese-English (Zh$\Rightarrow$En).
For De$\Rightarrow$En, we used a multi-domain dataset from \citet{aharoni2020unsupervised}.
To validate the generalization capability of the proposed method,
we set 4 seen domains (IT, Law, Medical and Subtitles) and 1 unseen domain (Koran).
Similarly, For Zh$\Rightarrow$En, we use UM-Corpus~\cite{tian2014corpus}, and also set 3 seen domains (Education, Spoken and Thesis) and 2 unseen domains (Science and Microblog).
For neuron-based methods, we randomly sampled 10k data for neuron selection and further sampled 2k data from this set for neuron-efficient fine-tuning.
And other baseline methods were fine-tuned on the full 10k data to ensure a fair comparison in terms of data volume.

\textbf{Backbone Models}
We take LLaMA2-7B-Chat\footnote{\url{https://huggingface.co/meta-llama/Llama-2-7b-chat}}, LLaMA3.1-8B-Instruct\footnote{\url{https://huggingface.co/meta-llama/Llama-3.1-8B-Instruct}}, and Qwen2.5-7B-Instruct\footnote{\url{https://huggingface.co/Qwen/Qwen2.5-7B-Instruct}}
as the backbone models for training.

\textbf{Implementation Details}
For neuron-based methods, we select 1\% neurons for fine-tuning.
And for our CANEFT, the threshold $\gamma$ (as defined in  Eq.~\ref{eq:8}) is dynamically determined by selecting the top $1\%$ of neurons with the highest MI scores across all domains as consensus-aligned neurons.
For LoRA-based baselines, including LoRA, LoRA-MLP, DoRA, we set rank to 8.
For LLaMA Pro, we add two adapter layers after layers 16 and 32.
All experiments are executed on 8 NVIDIA A40 GPUs.

\textbf{Baselines}
We compared our approach with several representative baselines.
\textbf{Base Inference} performs zero-shot inference using the original LLMs without any fine-tuning or domain adaptation.
\textbf{Full Fine-tuning} fine tunes all parameters of LLMs.
\textbf{LoRA} and \textbf{LoRA-MLP} (\textbf{L-MLP})~\cite{hu2022lora} apply low-rank adaptation to enable PEFT. While LoRA adds low-rank matrices to all modules and layers, LoRA-MLP restricts this to only the FFN modules, further reducing the number of trainable parameters.
\textbf{DoRA}~\cite{liu2024dora} decouples model parameters into magnitude and direction components, updating only the directional component to achieve efficient domain adaptation.
\textbf{LLaMA Pro} (\textbf{Pro})~\cite{wu2024llama} extends LLM depth by inserting identity-initialized transformer blocks, and fine-tuning only these additional layers using domain-specific data.
For neuron-based method, we adopt \textbf{LAPE}~\cite{tang2024language} as a representative baseline. 
This method leverages neuron activation probability entropy statistics to detect and fine-tune domain-specific neurons.
\textbf{Random Chosen Neurons} (\textbf{RCN}) fine-tunes a randomly selected subset of neurons, comparable in number to those selected by our CANEFT.

\subsection{Main Results}
Table~\ref{tab:translation-results} presents the performance of our method against several PEFT baselines across both seen and unseen domains, evaluated using \textit{SacreBLEU}\footnote{BLEU+case.mixed+numrefs.1+smooth.exp+tok.13a} (B) and COMET\footnote{https://huggingface.co/Unbabel/wmt22-cometkiwi-da} (C).

\textbf{Overall Performance}
Our CANEFT consistently delivers the best or second-best performance across all models and domains, underscoring its effectiveness in multi-domain translation. 
With Qwen2.5, it achieves 45.5 BLEU and 86.6 COMET on De$\Rightarrow$En and 29.6 BLEU and 83.1 COMET on Zh$\Rightarrow$En, surpassing the best-performing baseline by +1.2 BLEU and +0.7 COMET. 
Comparable gains are observed on LLaMA2 and LLaMA3.1,
where CANEFT achieves an average improves by +1.4-1.6 BLEU and +1-1.7 COMET on De$\Rightarrow$En, and +0.7-2 BLEU and +0.2-0.6 COMET on Zh$\Rightarrow$En over the strongest competing baselines.
Importantly, while full fine-tuning is the strongest baseline on average and achieves results comparable to CANEFT across most domains, CANEFT further surpasses it while updating only 1\% of the parameters.

CANEFT also demonstrated substantial gains on unseen domains. 
For example, in Zh$\Rightarrow$En, CANEFT showed enhanced robustness when handling linguistically diverse domains like Science (formal, technical) and Microblog (informal, colloquial). 
These domains represent significant textual distribution divergences from the training data, underscoring the enhanced robustness of our method in handling linguistic style variations and cross-domain adaptation.

\begin{figure*}[t]
    \includegraphics[width=0.48\textwidth]{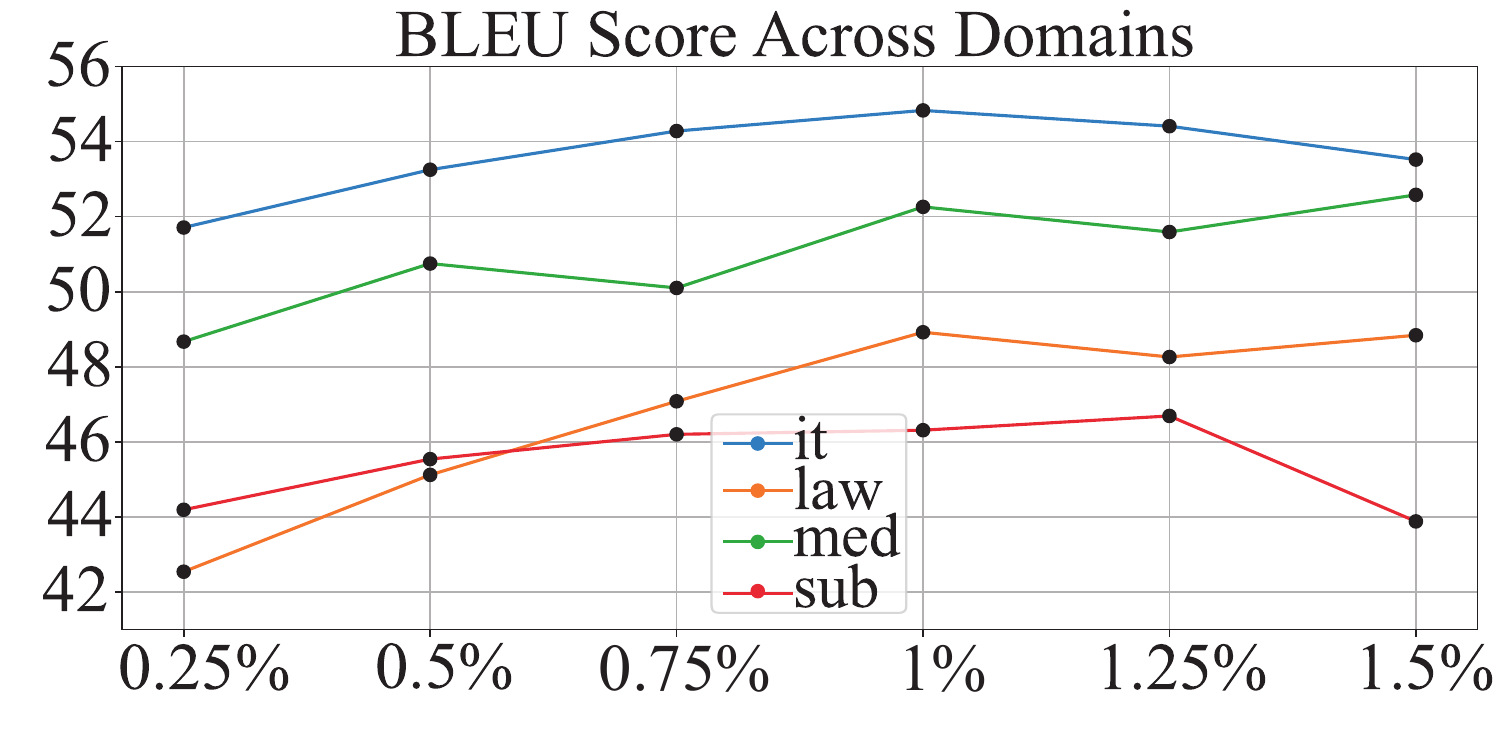}\hfill
    \includegraphics[width=0.48\textwidth]{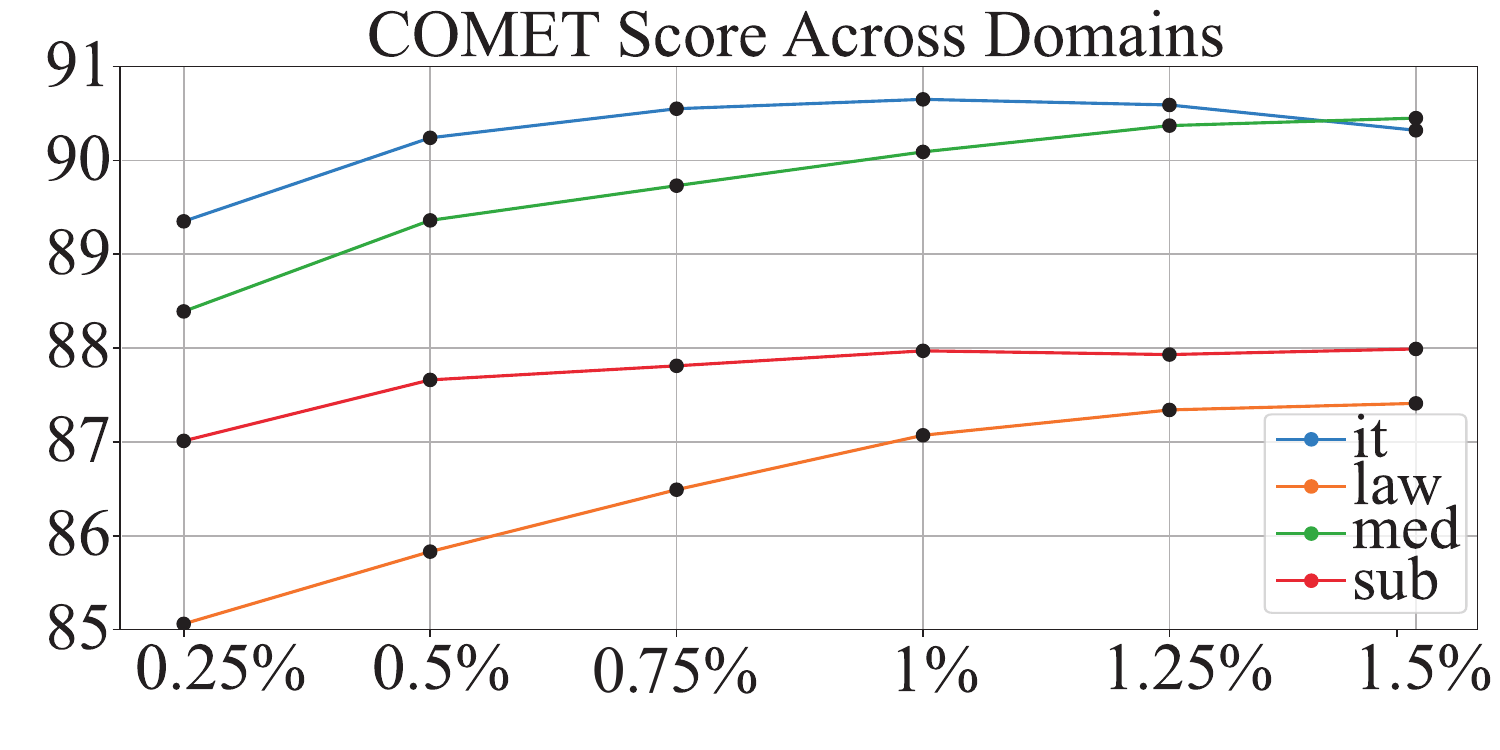}\hfill
    \caption{The impact of fine-tuning different multi-domain consensus-aligned neuron ratios on BLEU (left) and COMET (right) values with LLaMA3.1-8B-Instruct. In both plots, the x-axis shows neuron ratios, and the y-axis shows evaluation scores.}
    \label{fig:ratio-analysis}
\end{figure*}

Notably, LLaMA Pro slightly outperforms CANEFT in the Law domain (e.g., +0.8 BLEU on Qwen2.5), likely due to its 4 times additional parameters capturing legal-specific patterns. 
However, this comes at the expense of generalization, as LLaMA Pro suffers substantial drops in IT (-8.6 BLEU) and Subtitles (-9 BLEU) compared to Base Inference.
Similarly, most PEFT baselines show uneven domain gains, improving in some domains while degrading in others. This reflects parameter interference during multi-domain fine-tuning and underscores the risk of catastrophic forgetting.
In contrast, our method maintains robust performance across all domains, highlighting its advantage in balancing domain specialization with generalization.

Table~\ref{tab:translation-results} also shows that CANEFT introduces no additional parameters and updates only a minimal set of critical neurons. 
Although PEFT methods like LoRA and DoRA involve fewer trainable parameters, their performance across domains is significantly inferior, with some even underperforming base inference. 
In general, CANEFT strikes a superior balance between efficiency and translation quality.
Moreover, CANEFT significantly outperforms the RCN and LAPE across all domains, demonstrating that the selected neurons capture meaningful multi-domain consensus-aligned information rather than arbitrary features.
Furthermore, the consistent gains across 3 distinct backbones further demonstrate that our method is model-agnostic and applicable to various LLM-based translation frameworks.

\subsection{Ablation Study}
\textbf{\textit{w/o} MDMTN}: This variant computes MI between neurons and domain labels without identifying MDMT task-relevant neurons. 
As a result, MI is estimated over a larger and noisier neurons, making it harder to isolate meaningful consensus-aligned signals. 
As shown in Table~\ref{tab:ablation}, this leads to a noticeable performance drop across domains. 
These results underscore the importance of task-specific filtering prior to MI computation. 
In the absence of this filtering, the selected neurons include features not relevant to the task, which degrades the quality of adaptation.

\textbf{\textit{w/o} MDCAN}: This variant omits the multi-domain consensus-aligned neuron selection and directly fine-tunes the top 1\% of neurons based solely on importance scores, resulting in a clear performance drop. 
While MDMTN ensures task relevance, not all identified neurons could reach a consensus alignment across domains. Without MDCAN, fine-tuning a less-refined neuron set increasing risks of parameter interference across domains and dilutes domain-invariant signals. 
The MI-based MDCAN step is thus essential for identifying a harmonized neuron subset that enables robust and generalizable adaptation across domains.

\begin{table}[ht]
    \centering
    % \small
    \begin{tabular}{lcccc}
        \hline
        \textbf{Method} & \textbf{IT} & \textbf{Law} & \textbf{Medical} & \textbf{Subtitles} \\
        \hline
        CANEFT & 54.8 & 50.9 & 52.3 & 46.3 \\
        \textit{w/o} MDMTN & 48.4 & 44.2 & 47.3 & 40.2 \\
        \textit{w/o} MDCAN & 46.2 & 40.6 & 44.9 & 41.8 \\
    \bottomrule
    \end{tabular}
    \caption{This table show ablation study and report BLEU scores in De$\Rightarrow$En with LLaMA3.1-8B-Instruct.}
    \label{tab:ablation}
\end{table}

\begin{figure*}[t]
    \centering
    \includegraphics[width=1\linewidth]{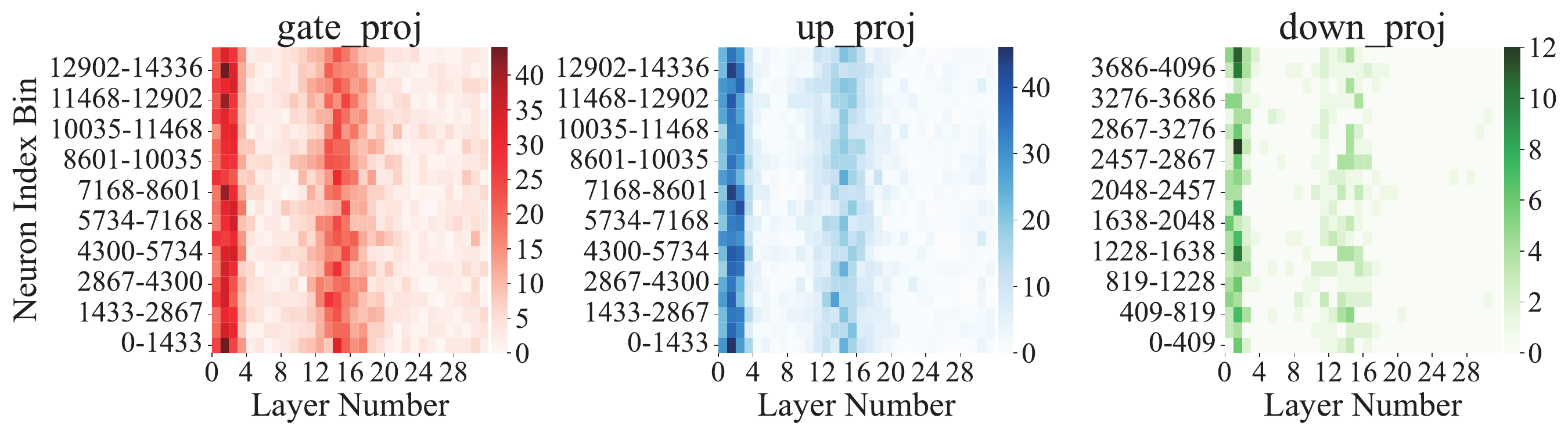}
    \caption{Distribution of multi-domain consensus-aligned neurons across layers within the FFN's \textit{gate\_proj}, \textit{up\_proj} and \textit{down\_proj} modules of LLaMA3.1-8B-Instruct. In each plots, the x-axis denotes the layer number, and the y-axis corresponds to neuron index bins, derived by segmenting the full range of neuron indices into 15 divisions.}
    \label{fig:neuron-distribution}
\end{figure*}

\begin{figure}[t]
    \centering
    \includegraphics[width=1\linewidth]{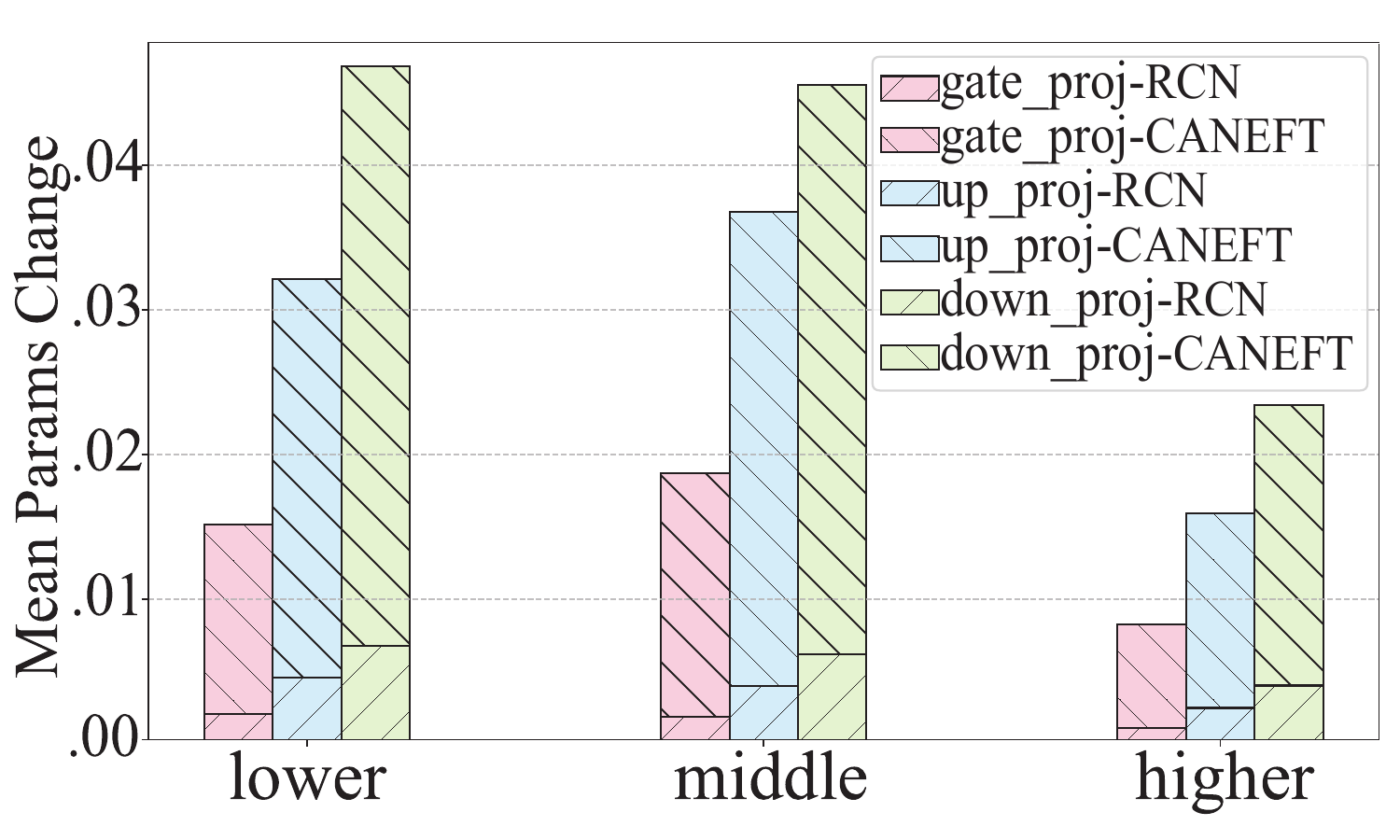}
    \caption{Gradient changes in consensus-aligned neurons (CANEFT) and randomly chosen neurons (RCN) within the FFN's \textit{gate\_proj}, \textit{up\_proj} and \textit{down\_proj} modules of LLaMA3.1-8B-Instruct. Layers are grouped by depth into lower, middle, and higher sections. In each bar, the lower segment represents the gradient changes of randomly chosen neurons, while the upper segment corresponds to those of consensus-aligned neurons.}
    \label{fig:gradient-changes-analysis}
\end{figure}

\subsection{Impact of different neuron ratio}
We investigate how varying the proportion of selected multi-domain consensus-aligned neurons (from 0.25\% to 1.5\%) influences performance using LLaMA3.1-8B-Instruct.

As shown in Figure~\ref{fig:ratio-analysis}, increasing the ratio of selected neurons consistently improves translation quality across both BLEU and COMET metrics.
Notably, the IT and Law domains exhibit the most substantial improvements, with BLEU gains of 3–6 points, suggesting that incorporating more consensus-aligned neurons facilitates more effective cross-domain knowledge transfer. 
COMET scores follow a similar upward trend, reflecting enhanced semantic adequacy and fluency.
However, performance plateaus beyond 0.75\% and even degrades when over 1.25\% in most domains, suggesting that most informative neurons have already been leveraged, and additional parameters contribute marginal returns. 
These results confirm that fine-tuning only 0.5\%–1\% of well-chosen neurons is sufficient to achieve strong multi-domain translation performance while avoiding unnecessary parameter updates.

\subsection{Gradient changes between randomly selected and consensus-aligned neurons}
To assess the relevance of the selected consensus-aligned neurons in MDMT, we analyze their behavior from a gradient perspective.
We compare the mean absolute parameter changes in LLaMA3.1-8B’s FFN components (\textit{gate\_proj}, \textit{up\_proj}, \textit{down\_proj}) under two neuron selection strategies: random selection and multi-domain consensus-aligned neuron selection. 
Model layers are grouped into three ranges: lower (layers 0–10), middle (layers 11–20), and higher (layers 21–32). As shown in Figure~\ref{fig:gradient-changes-analysis}, consensus-aligned neurons consistently exhibit larger gradient updates across all components and layers.

These substantial gradient differentials suggest that consensus-aligned neurons possess higher optimization potential for MDMT. 
Our selection method effectively identifies parameters most responsive to multi-domain adaptation, enabling the LLM to refine these neurons for improved translation performance. Such neurons undergo more meaningful, task-specific transformations, whereas randomly selected neurons display weaker and less focused changes. Moreover, the pronounced shifts in \textit{up\_proj} and \textit{down\_proj} highlight their pivotal role in domain adaptation,
underscoring their importance in facilitating deeper integration of domain knowledge into the model’s representations.

\subsection{Distribution of consensus-aligned neurons}
To visually demonstrate the distribution of multi-domain consensus-aligned neurons across different layers in LLaMA3.1-8B-Instruct. 
Figure~\ref{fig:neuron-distribution} provides heatmaps of the selected neurons across the 32 layers for each of the three FFN projections.
The selected neurons are unevenly distributed, with higher density in lower and middle layers particularly in \textit{gate\_proj} and \textit{up\_proj}. 
In contrast, \textit{down\_proj} neurons are selected sparsely, indicating a relatively limited role in cross-domain generalization. 
This pattern aligns with the understanding that lower layers capture general syntactic and semantic features beneficial for transfer, while higher layers encode more domain-specific information thus contributing fewer consensus-aligned neurons.

The salience of \textit{gate\_proj} and \textit{up\_proj} points to their critical role in regulating information flow and enriching representations for multi-domain processing.
\textit{gate\_proj} modulates the activation gate, while \textit{up\_proj} expands hidden representations. 
The high density of selected neurons in these components suggests that consensus knowledge are more effectively encoded and propagated during the gating and expansion phases, rather than in the dimensionality reduction stage (\textit{down\_proj}).
This is further supported by Figure~\ref{fig:gradient-changes-analysis}, where these components show stronger fine-tuning signals, indicating their structural and functional significance in encoding transferable knowledge.

\section{Conclusion}
We propose a neuron-efficient fine-tuning framework for MDMT that selectively updates consensus-aligned neurons, identified by maximizing MI between neuron behavior and domain features. This method improves translation quality and mitigates parameter interference and domain-specific overfitting. Unlike existing PEFT methods that require additional parameters, our method generalizes well to unseen domains with no extra parameters, and achieves SOTA performance across 10 domains on 3 LLMs, highlighting the promise of neuronal domain adaptation in LLMs.

\section{Ethics Statement}
This research adheres to a strict ethical framework as it does not involve any ethical issues. The data constructed for this research is derived solely from open-source data, and the large language models employed in this study follows their declared licenses. I have fully informed the participants of all instructions, to ensure they are fully aware and consenting to participate in this work.

\section{Acknowledgments}
This research was supported by the National Natural Science Foundation of China (Grant Nos. U24A20334, 62366027, U21B2027, 62266027), the Yunnan Provincial Major Science and Technology Special Plan Projects (Grant Nos. 202303AP140008, 202203AA080004, 202302AD080003, 202401BC070021), the General Projects of Basic Research in Yunnan Province (Grant Nos. 202201BE070001-021).

% \bigskip
% \noindent Thank you for reading these instructions carefully. We look forward to receiving your electronic files!

\bibliography{aaai2026}

@inproceedings{mao-etal-2025-multilingual,
    title = "Multilingual Knowledge Graph Completion via Efficient Multilingual Knowledge Sharing",
    author = "Mao, Cunli  and
      Gao, Xiaofei  and
      Song, Ran  and
      He, Shizhu  and
      Gao, Shengxiang  and
      Liu, Kang  and
      Yu, Zhengtao",
    editor = "Christodoulopoulos, Christos  and
      Chakraborty, Tanmoy  and
      Rose, Carolyn  and
      Peng, Violet",
    booktitle = "Findings of the Association for Computational Linguistics: EMNLP 2025",
    month = nov,
    year = "2025",
    address = "Suzhou, China",
    publisher = "Association for Computational Linguistics",
    url = "https://aclanthology.org/2025.findings-emnlp.577/",
    doi = "10.18653/v1/2025.findings-emnlp.577",
    pages = "10882--10896",
    ISBN = "979-8-89176-335-7"
}

@article{pang-etal-2025-salute,
    title = "Salute the Classic: Revisiting Challenges of Machine Translation in the Age of Large Language Models",
    author = "Pang, Jianhui  and
      Ye, Fanghua  and
      Wong, Derek Fai  and
      Yu, Dian  and
      Shi, Shuming  and
      Tu, Zhaopeng  and
      Wang, Longyue",
    journal = "Transactions of the Association for Computational Linguistics",
    volume = "13",
    year = "2025",
    address = "Cambridge, MA",
    publisher = "MIT Press",
    url = "https://aclanthology.org/2025.tacl-1.4/",
    doi = "10.1162/tacl_a_00730",
    pages = "73--95",
}

@article{he-etal-2024-exploring,
    title = "Exploring Human-Like Translation Strategy with Large Language Models",
    author = "He, Zhiwei  and
      Liang, Tian  and
      Jiao, Wenxiang  and
      Zhang, Zhuosheng  and
      Yang, Yujiu  and
      Wang, Rui  and
      Tu, Zhaopeng  and
      Shi, Shuming  and
      Wang, Xing",
    journal = "Transactions of the Association for Computational Linguistics",
    volume = "12",
    year = "2024",
    address = "Cambridge, MA",
    publisher = "MIT Press",
    url = "https://aclanthology.org/2024.tacl-1.13/",
    doi = "10.1162/tacl_a_00642",
    pages = "229--246",
}

@inproceedings{molchanov2017pruning,
  title={Pruning Convolutional Neural Networks for Resource Efficient Inference},
  author={Molchanov, Pavlo and Tyree, Stephen and Karras, Tero and Aila, Timo and Kautz, Jan},
  booktitle={International Conference on Learning Representations},
  year={2017},
  url={https://doi.org/10.48550/arXiv.2107.06569}
}

@article{Sievers2024consensus,
  author  = {Sievers, Beau and Welker, Christopher and Hasson, Uri and Kleinbaum, Adam M. and Wheatley, Thalia},
  title   = {Consensus-building conversation leads to neural alignment},
  journal = {Nature Communications},
  year    = {2024},
  date    = {2024-05-10},
  volume  = {15},
  number  = {1},
  pages   = {3936},
  doi     = {10.1038/s41467-023-43253-8},
  url     = {https://doi.org/10.1038/s41467-023-43253-8}
}

@article{li2024knowledge,
  title={Knowledge Editing for Large Language Model with Knowledge Neuronal Ensemble},
  author={Li, Yongchang and Zhu, Yujin and Yan, Tao and Fan, Shijian and Wu, Gang and Xu, Liang},
  journal={arXiv preprint arXiv:2412.20637},
  year={2024},
  url={https://arxiv.org/abs/2412.20637}
}

@article{jiang2024neuron,
  title={Neuron-level sequential editing for large language models},
  author={Jiang, Houcheng and Fang, Junfeng and Zhang, Tianyu and Zhang, An and Wang, Ruipeng and Liang, Tao and Wang, Xiang},
  journal={arXiv preprint arXiv:2410.04045},
  year={2024},
  url={https://arxiv.org/abs/2410.04045},
}

@article{hersche2024towards,
  title={Towards Learning to Reason: Comparing LLMs with Neuro-Symbolic on Arithmetic Relations in Abstract Reasoning},
  author={Hersche, Michael and Camposampiero, Giacomo and Wattenhofer, Roger and Sebastian, Abu and Rahimi, Abbas},
  journal={arXiv preprint arXiv:2412.05586},
  year={2024},
  url={https://arxiv.org/abs/2412.05586},
}

@inproceedings{rai2024investigation,
  title={An Investigation of Neuron Activation as a Unified Lens to Explain Chain-of-Thought Eliciting Arithmetic Reasoning of LLMs},
  author={Rai, Daking and Yao, Ziyu},
  booktitle={Proceedings of the 62nd Annual Meeting of the Association for Computational Linguistics (Volume 1: Long Papers)},
  pages={7174--7193},
  year={2024},
  url={https://aclanthology.org/2024.acl-long.387.pdf},
}

@inproceedings{hu-etal-2023-llm,
    title = "{LLM}-Adapters: An Adapter Family for Parameter-Efficient Fine-Tuning of Large Language Models",
    author = "Hu, Zhiqiang  and
      Wang, Lei  and
      Lan, Yihuai  and
      Xu, Wanyu  and
      Lim, Ee-Peng  and
      Bing, Lidong  and
      Xu, Xing  and
      Poria, Soujanya  and
      Lee, Roy",
    editor = "Bouamor, Houda  and
      Pino, Juan  and
      Bali, Kalika",
    booktitle = "Proceedings of the 2023 Conference on Empirical Methods in Natural Language Processing",
    month = dec,
    year = "2023",
    address = "Singapore",
    publisher = "Association for Computational Linguistics",
    url = "https://aclanthology.org/2023.emnlp-main.319/",
    doi = "10.18653/v1/2023.emnlp-main.319",
    pages = "5254--5276",
}

@inproceedings{eschbach2024exploring,
  title={Exploring the effectiveness of LLM domain adaptation for business it machine translation},
  author={Eschbach-Dymanus, Johannes and Essenberger, Frank and Buschbeck, Bianka and Exel, Miriam},
  booktitle={Proceedings of the 25th Annual Conference of the European Association for Machine Translation (Volume 1)},
  pages={610--622},
  year={2024},
  url={https://aclanthology.org/2024.eamt-1.51/},
}

@inproceedings{vilar2023prompting,
  title={Prompting PaLM for Translation: Assessing Strategies and Performance},
  author={Vilar, David and Freitag, Markus and Cherry, Colin and Luo, Jiaming and Ratnakar, Viresh and Foster, George},
  booktitle={Proceedings of the 61st Annual Meeting of the Association for Computational Linguistics (Volume 1: Long Papers)},
  pages={15406--15427},
  year={2023},
  url={https://aclanthology.org/2023.acl-long.859.pdf},
}

@article{huang2024survey,
  title={A survey on large language models with multilingualism: Recent advances and new frontiers},
  author={Huang, Kaiyu and Mo, Fengran and Zhang, Xinyu and Li, Hongliang and Li, You and Zhang, Yuanchi and Yi, Weijian and Mao, Yulong and Liu, Jinchen and Xu, Yuzhuang and others},
  journal={arXiv preprint arXiv:2405.10936},
  year={2024},
  url={https://arxiv.org/abs/2405.10936},
}

@article{zhao2023survey,
  title={A survey of large language models},
  author={Zhao, Wayne Xin and Zhou, Kun and Li, Junyi and Tang, Tianyi and Wang, Xiaolei and Hou, Yupeng and Min, Yingqian and Zhang, Beichen and Zhang, Junjie and Dong, Zican and others},
  journal={arXiv preprint arXiv:2303.18223},
  year={2023},
  url={https://arxiv.org/abs/2303.18223},
}

@article{Li_Wang_Yu_2020, 
    title={MetaMT, a Meta Learning Method Leveraging Multiple Domain Data for Low Resource Machine Translation}, 
    volume={34}, 
    url={https://ojs.aaai.org/index.php/AAAI/article/view/6339}, 
    DOI={10.1609/aaai.v34i05.6339}, 
    number={05}, 
    journal={Proceedings of the AAAI Conference on Artificial Intelligence}, 
    author={Li, Rumeng and Wang, Xun and Yu, Hong}, year={2020}, 
    month={Apr.}, 
    pages={8245-8252} 
}

@inproceedings{moslem2023adaptive,
  title={Adaptive Machine Translation with Large Language Models},
  author={Moslem, Yasmin and Haque, Rejwanul and Kelleher, John and Way, Andy},
  booktitle={Proceedings of the 24th Annual Conference of the European Association for Machine Translation},
  pages={227--237},
  year={2023},
  url={https://aclanthology.org/2023.eamt-1.22/},
}

@article{saunders2022domain,
  title={Domain adaptation and multi-domain adaptation for neural machine translation: A survey},
  author={Saunders, Danielle},
  journal={Journal of Artificial Intelligence Research},
  volume={75},
  pages={351--424},
  year={2022},
  url={https://doi.org/10.1613/jair.1.13566},
}

@article{pham2021revisiting,
  title={Revisiting multi-domain machine translation},
  author={Pham, MinhQuang and Crego, Josep Maria and Yvon, Fran{\c{c}}ois},
  journal={Transactions of the Association for Computational Linguistics},
  volume={9},
  pages={17--35},
  year={2021},
  publisher={MIT Press One Rogers Street, Cambridge, MA 02142-1209, USA journals-info~…},
  url={https://aclanthology.org/2021.tacl-1.2/},
}

@article{ghazvininejad2023dictionary,
  title={Dictionary-based phrase-level prompting of large language models for machine translation},
  author={Ghazvininejad, Marjan and Gonen, Hila and Zettlemoyer, Luke},
  journal={arXiv preprint arXiv:2302.07856},
  year={2023},
  url={https://arxiv.org/abs/2302.07856},
}

@inproceedings{alves-etal-2023-steering,
    title = "Steering Large Language Models for Machine Translation with Finetuning and In-Context Learning",
    author = "Alves, Duarte  and
      Guerreiro, Nuno  and
      Alves, Jo{\~a}o  and
      Pombal, Jos{\'e}  and
      Rei, Ricardo  and
      de Souza, Jos{\'e}  and
      Colombo, Pierre  and
      Martins, Andre",
    editor = "Bouamor, Houda  and
      Pino, Juan  and
      Bali, Kalika",
    booktitle = "Findings of the Association for Computational Linguistics: EMNLP 2023",
    month = dec,
    year = "2023",
    address = "Singapore",
    publisher = "Association for Computational Linguistics",
    url = "https://aclanthology.org/2023.findings-emnlp.744/",
    doi = "10.18653/v1/2023.findings-emnlp.744",
    pages = "11127--11148",
}

@article{zheng2024fine,
  title={Fine-tuning large language models for domain-specific machine translation},
  author={Zheng, Jiawei and Hong, Hanghai and Liu, Feiyan and Wang, Xiaoli and Su, Jingsong and Liang, Yonggui and Wu, Shikai},
  journal={arXiv preprint arXiv:2402.15061},
  year={2024},
  url={https://arxiv.org/abs/2402.15061},
}

@article{dragft,
    author = {Zheng, Jiawei and Hong, Hanghai and Liu, Feiyan and Wang, Xiaoli and Su, Jingsong},
    title = {DragFT: Adapting Large Language Models with Dictionary and Retrieval Augmented Fine-tuning for Domain-specific Machine Translation} ,
    journal = {arXiv preprint arXiv:arXiv:2402.15061v2},
    year = {2024},
    url = {https://arxiv.org/abs/2402.15061v2},
}

@inproceedings{hu2024large,
  title={Large Language Model for Multi-Domain Translation: Benchmarking and Domain CoT Fine-tuning},
  author={Hu, Tianxiang and Zhang, Pei and Yang, Baosong and Xie, Jun and Wong, Derek and Wang, Rui},
  booktitle={Findings of the Association for Computational Linguistics: EMNLP 2024},
  pages={5726--5746},
  year={2024},
  url={https://aclanthology.org/2024.findings-emnlp.328.pdf},
}

@article{li2025leveraging,
  title={Leveraging Domain Knowledge at Inference Time for LLM Translation: Retrieval versus Generation},
  author={Li, Bryan and Luo, Jiaming and Briakou, Eleftheria and Cherry, Colin},
  journal={arXiv preprint arXiv:2503.05010},
  year={2025},
  url={https://arxiv.org/abs/2503.05010},
}

@inproceedings{aycock2024topic,
  title={Topic-guided example selection for domain adaptation in llm-based machine translation},
  author={Aycock, Seth and Bawden, Rachel},
  booktitle={Proceedings of the 18th Conference of the European Chapter of the Association for Computational Linguistics: Student Research Workshop},
  pages={175--195},
  year={2024},
  url={https://aclanthology.org/2024.eacl-srw.13/},
}

@inproceedings{niudoes,
  title={What does the Knowledge Neuron Thesis Have to do with Knowledge?},
  author={Niu, Jingcheng and Liu, Andrew and Zhu, Zining and Penn, Gerald},
  booktitle={The Twelfth International Conference on Learning Representations},
  year={2024},
  url={https://arxiv.org/abs/2405.02421},
}

@inproceedings{dai2022knowledge,
  title={Knowledge Neurons in Pretrained Transformers},
  author={Dai, Damai and Dong, Li and Hao, Yaru and Sui, Zhifang and Chang, Baobao and Wei, Furu},
  booktitle={Proceedings of the 60th Annual Meeting of the Association for Computational Linguistics (Volume 1: Long Papers)},
  pages={8493--8502},
  year={2022},
  url={https://aclanthology.org/2022.acl-long.581.pdf},
}

@inproceedings{tan2024neuron,
  title={Neuron Specialization: Leveraging Intrinsic Task Modularity for Multilingual Machine Translation},
  author={Tan, Shaomu and Wu, Di and Monz, Christof},
  booktitle={Proceedings of the 2024 Conference on Empirical Methods in Natural Language Processing},
  pages={6506--6527},
  year={2024},
  url={https://aclanthology.org/2024.emnlp-main.374/},
}

@inproceedings{voita2024neurons,
  title={Neurons in Large Language Models: Dead, N-gram, Positional},
  author={Voita, Elena and Ferrando, Javier and Nalmpantis, Christoforos},
  booktitle={Findings of the Association for Computational Linguistics ACL 2024},
  pages={1288--1301},
  year={2024},
  url={https://aclanthology.org/2024.findings-acl.75.pdf},
}

@inproceedings{chen2024journey,
  title={Journey to the center of the knowledge neurons: Discoveries of language-independent knowledge neurons and degenerate knowledge neurons},
  author={Chen, Yuheng and Cao, Pengfei and Chen, Yubo and Liu, Kang and Zhao, Jun},
  booktitle={Proceedings of the AAAI Conference on Artificial Intelligence},
  volume={38},
  pages={17817--17825},
  year={2024},
  url={https://doi.org/10.1609/aaai.v38i16.29735},
}

@inproceedings{xie2021importance,
  title={Importance-based Neuron Allocation for Multilingual Neural Machine Translation},
  author={Xie, Wanying and Feng, Yang and Gu, Shuhao and Yu, Dong},
  booktitle={Proceedings of the 59th Annual Meeting of the Association for Computational Linguistics and the 11th International Joint Conference on Natural Language Processing (Volume 1: Long Papers)},
  pages={5725--5737},
  year={2021},
  url={https://aclanthology.org/2021.acl-long.445.pdf},
}

@inproceedings{leng2025towards,
  title={Towards Understanding Multi-Task Learning (Generalization) of LLMs via Detecting and Exploring Task-Specific Neurons},
  author={Leng, Yongqi and Xiong, Deyi},
  booktitle={Proceedings of the 31st International Conference on Computational Linguistics},
  pages={2969--2987},
  year={2025},
  url={https://aclanthology.org/2025.coling-main.200.pdf},
}

@inproceedings{xu2025let,
  title={Let’s Focus on Neuron: Neuron-Level Supervised Fine-tuning for Large Language Model},
  author={Xu, Haoyun and Zhan, Runzhe and Ma, Yingpeng and Wong, Derek F and Chao, Lidia S},
  booktitle={Proceedings of the 31st International Conference on Computational Linguistics},
  pages={9393--9406},
  year={2025},
  url={https://aclanthology.org/2025.coling-main.630.pdf},
}

@inproceedings{song2024does,
  title={Does Large Language Model Contain Task-Specific Neurons?},
  author={Song, Ran and He, Shizhu and Jiang, Shuting and Xian, Yantuan and Gao, Shengxiang and Liu, Kang and Yu, Zhengtao},
  booktitle={Proceedings of the 2024 Conference on Empirical Methods in Natural Language Processing},
  pages={7101--7113},
  year={2024},
  url={https://aclanthology.org/2024.emnlp-main.403/},
}

@inproceedings{tang2024language,
  title={Language-Specific Neurons: The Key to Multilingual Capabilities in Large Language Models},
  author={Tang, Tianyi and Luo, Wenyang and Huang, Haoyang and Zhang, Dongdong and Wang, Xiaolei and Zhao, Wayne Xin and Wei, Furu and Wen, Ji-Rong},
  booktitle={Proceedings of the 62nd Annual Meeting of the Association for Computational Linguistics (Volume 1: Long Papers)},
  pages={5701--5715},
  year={2024},
  url={https://aclanthology.org/2024.acl-long.309/},
}

@inproceedings{zhu2024landermt,
  title={LANDeRMT: Dectecting and Routing Language-Aware Neurons for Selectively Finetuning LLMs to Machine Translation},
  author={Zhu, Shaolin and Pan, Leiyu and Li, Bo and Xiong, Deyi},
  booktitle={Proceedings of the 62nd Annual Meeting of the Association for Computational Linguistics (Volume 1: Long Papers)},
  pages={12135--12148},
  year={2024},
  url={https://aclanthology.org/2024.acl-long.656/},
}

@article{hu2022lora,
  title={Lora: Low-rank adaptation of large language models},
  author={Hu, Edward J and Shen, Yelong and Wallis, Phillip and Allen-Zhu, Zeyuan and Li, Yuanzhi and Wang, Shean and Wang, Lu and Chen, Weizhu and others},
  journal={ICLR},
  volume={1},
  number={2},
  pages={3},
  year={2022},
  url={https://arxiv.org/abs/2106.09685}
}

@inproceedings{liu2024dora,
  title={Dora: Weight-decomposed low-rank adaptation},
  author={Liu, Shih-Yang and Wang, Chien-Yi and Yin, Hongxu and Molchanov, Pavlo and Wang, Yu-Chiang Frank and Cheng, Kwang-Ting and Chen, Min-Hung},
  booktitle={Forty-first International Conference on Machine Learning},
  year={2024},
  url={https://arxiv.org/abs/2402.09353}
}

@inproceedings{wu2024llama,
    title = "{LL}a{MA} Pro: Progressive {LL}a{MA} with Block Expansion",
    author = "Wu, Chengyue  and
      Gan, Yukang  and
      Ge, Yixiao  and
      Lu, Zeyu  and
      Wang, Jiahao  and
      Feng, Ye  and
      Shan, Ying  and
      Luo, Ping",
    editor = "Ku, Lun-Wei  and
      Martins, Andre  and
      Srikumar, Vivek",
    booktitle = "Proceedings of the 62nd Annual Meeting of the Association for Computational Linguistics (Volume 1: Long Papers)",
    month = aug,
    year = "2024",
    address = "Bangkok, Thailand",
    publisher = "Association for Computational Linguistics",
    url = "https://aclanthology.org/2024.acl-long.352/",
    doi = "10.18653/v1/2024.acl-long.352",
    pages = "6518--6537",
}

@inproceedings{aharoni2020unsupervised,
    title = "Unsupervised Domain Clusters in Pretrained Language Models",
    author = "Aharoni, Roee  and
      Goldberg, Yoav",
    editor = "Jurafsky, Dan  and
      Chai, Joyce  and
      Schluter, Natalie  and
      Tetreault, Joel",
    booktitle = "Proceedings of the 58th Annual Meeting of the Association for Computational Linguistics",
    month = jul,
    year = "2020",
    address = "Online",
    publisher = "Association for Computational Linguistics",
    url = "https://aclanthology.org/2020.acl-main.692/",
    doi = "10.18653/v1/2020.acl-main.692",
    pages = "7747--7763",
}

@inproceedings{tian2014corpus,
    title = "{UM}-Corpus: A Large {E}nglish-{C}hinese Parallel Corpus for Statistical Machine Translation",
    author = "Tian, Liang  and
      Wong, Derek F.  and
      Chao, Lidia S.  and
      Quaresma, Paulo  and
      Oliveira, Francisco  and
      Lu, Yi  and
      Li, Shuo  and
      Wang, Yiming  and
      Wang, Longyue",
    editor = "Calzolari, Nicoletta  and
      Choukri, Khalid  and
      Declerck, Thierry  and
      Loftsson, Hrafn  and
      Maegaard, Bente  and
      Mariani, Joseph  and
      Moreno, Asuncion  and
      Odijk, Jan  and
      Piperidis, Stelios",
    booktitle = "Proceedings of the Ninth International Conference on Language Resources and Evaluation ({LREC}`14)",
    month = may,
    year = "2014",
    address = "Reykjavik, Iceland",
    publisher = "European Language Resources Association (ELRA)",
    url = "https://aclanthology.org/L14-1604/",
    pages = "1837--1842",
}

\end{document}